\documentclass[conference]{IEEEtran}
\usepackage{cite}
\usepackage{url}
\usepackage{amsmath,amssymb,amsfonts}
\usepackage{algorithmic}
\usepackage{graphicx}
\usepackage{textcomp}
\usepackage{xcolor}
\usepackage{graphicx}
\usepackage{amsmath}
\usepackage{multirow}
\usepackage{array}
\usepackage{amssymb,tikz-cd, adjustbox, graphicx}
\usetikzlibrary{arrows.meta, positioning}
\usepackage{tikz}
\usepackage{makecell}
\usepackage[switch]{lineno}

\def\BibTeX{{\rm B\kern-.05em{\sc i\kern-.025em b}\kern-.08em
    T\kern-.1667em\lower.7ex\hbox{E}\kern-.125emX}}

\newcommand{\compacttables}{
  \footnotesize
  \setlength{\tabcolsep}{5pt} % reduce column spacing
  \renewcommand{\arraystretch}{1.152} % reduce row spacing
}

\begin{document}

\title{From Roots to Rewards: Dynamic Tree Reasoning with Reinforcement Learning}

\author{
\IEEEauthorblockN{Ahmed Bahloul, Simon Malberg}
\IEEEauthorblockA{\textit{School of Computation, Information and Technology} \\
\textit{Technical University of Munich}\\
Munich, Germany \\
\{ahmed.bahloul, simon.malberg\}@tum.de}
}
\maketitle

\begin{abstract}
Modern language models address complex questions through chain-of-thought (CoT) reasoning and retrieval augmentation, yet struggle with error propagation and knowledge integration. Tree-structured reasoning methods, particularly the Probabilistic Tree-of-Thought (ProbTree) framework, mitigate these issues by decomposing questions into hierarchical structures and selecting answers through confidence-weighted aggregation of parametric and retrieved knowledge. However, ProbTree's static implementation introduces two key limitations: (1) the reasoning tree is fixed during the initial construction phase, preventing dynamic adaptation to intermediate results, and (2) each node requires exhaustive evaluation of all possible solution strategies, creating computational inefficiency.

We present a dynamic reinforcement learning framework that transforms tree-based reasoning into an adaptive process. Our approach incrementally constructs the reasoning tree based on real-time confidence estimates, while learning optimal policies for action selection (decomposition, retrieval, or aggregation). This maintains ProbTree's probabilistic rigor while improving both solution quality and computational efficiency through selective expansion and focused resource allocation.

The work establishes a new paradigm for tree-structured reasoning that balances the reliability of probabilistic frameworks with the flexibility required for real-world question answering systems.
\end{abstract}

\section{Introduction}

Large language models (LLMs) augmented with tree-structured reasoning, such as \textit{ProbTree}~\cite{probtree}, have shown promise in answering knowledge-intensive complex questions by decomposing queries into sub-questions and probabilistically combining parametric and retrieved knowledge~\cite{llm_as_kb}.

The ProbTree framework tackles complex question answering by constructing a tree-structured reasoning process. At the root of the tree, the input question is decomposed into smaller sub-questions, forming hierarchical nodes in the reasoning tree. At each node, the framework employs three possible solution strategies: (1) Closed-Book (CB), which leverages parametric knowledge from the language model alone; (2) Open-Book (OB), relying on external retrieval~\cite{rag} to augment the language model’s outputs; and (3) Child-aggregation (CA), where answers are recursively constructed by combining the outputs of child nodes in the tree. Each node computes a probabilistic confidence score for the outputs of these strategies, and the highest-confidence answer is selected. This selection process ensures that responses are grounded in both parametric (language model) and non-parametric (retrieved) knowledge, making ProbTree well-suited for handling knowledge-intensive queries. However, this static pipeline—where the full reasoning tree and strategy choices are decided in a single step—limits flexibility and introduces computational inefficiencies. Moreover, its brute-force strategy of evaluating all possible solutions per node (Closed-Book, Open-Book, and Child-aggregation) introduces computational inefficiency, especially for deep or wide trees.\footnote{Code available at: \url{https://github.com/ahmedehabb/From-Roots-to-Rewards-Dynamic-Tree-Reasoning-with-RL}}

\subsection{Motivation}
The original ProbTree framework faces two critical challenges:

\begin{itemize}
    \item \textbf{Static Tree Construction}: The query tree is generated once during an initial understanding phase and remains fixed throughout reasoning. This rigidity prevents the system from recovering from poor initial decompositions or adapting to new information discovered during reasoning.
    
    \item \textbf{Computational Overhead}: For each node, the model evaluates all three reasoning strategies (Closed-Book, Open-Book, and Child-aggregation) before selecting the highest-confidence answer. This exhaustive approach becomes prohibitively expensive for complex questions with deep decomposition trees.
\end{itemize}

\subsection{Our Approach}
We propose a dynamic reinforcement learning (RL) framework that addresses these limitations through:

\begin{itemize}
    \item \textbf{On-Demand Tree Construction}: The query tree is built incrementally during reasoning, with child nodes expanded only when necessary (e.g., when confidence in the current decomposition is low).
    
    \item \textbf{Adaptive Action Selection}: An RL agent learns to choose the most promising reasoning strategy at each step, eliminating the need to evaluate all possibilities exhaustively.
\end{itemize}

Our method maintains ProbTree's probabilistic confidence mechanism while introducing greater flexibility and efficiency. For example, the agent can dynamically decide to:
\begin{itemize}
    \item Decompose a child node further if the current sub-question remains too complex
    \item Switch to retrieval-augmented reasoning when parametric knowledge is insufficient
    \item Introduce new actions, such as \textbf{Reformulation} and \textbf{Resampling}, to enhance the expressiveness and adaptability of the reasoning process.
\end{itemize}

\subsection{Contributions}
The key contributions of this work include:
\begin{itemize}
    \item A novel \textit{dynamic tree-of-thought} framework where decomposition and reasoning are interleaved and guided by RL policies
    \item Elimination of brute-force evaluation through learned action selection strategies
    \item Empirical validation showing competitive performance with reduced computational cost on standard benchmarks (HotpotQA~\cite{hotpotqa}, MusiQue~\cite{musique}, 2WikiMultihopQA~\cite{2wiki})
\end{itemize}

This work bridges the gap between rigid hierarchical reasoning and adaptive problem-solving, offering a more scalable paradigm for complex QA systems. The dynamic nature of our approach makes it particularly suitable for real-world applications where question complexity and available knowledge may vary significantly.

\section{Reinforcement Learning Framework for Dynamic Tree-of-Thought Reasoning}

To overcome the limitations of static and brute-force tree generation in the original ProbTree framework~\cite{probtree}, we introduce a RL based agent that dynamically generates reasoning trees and selects actions at each step to answer knowledge-intensive questions efficiently and accurately.

\subsection{Problem Formulation}
We formulate the decision process of tree construction and answer selection as a Markov Decision Process (MDP)~\cite{rl_intro}, where:
\begin{itemize}
    \item \textbf{State ($s$)}: Encodes information about the current node in the tree, including semantic features, structural position, confidence scores, and answer embeddings~\cite{sentence_bert}.
    \item \textbf{Action ($a$)}: One of several reasoning strategies including Closed-Book (CB), Open-Book (OB), Child Aggregation (Child), and their reformulated or resampling variants.
    \item \textbf{Reward ($r$)}: A combination of semantic accuracy of the answer and a penalty for computational cost, specifically the number of LLM calls made. This design is inspired by cost-aware RL frameworks~\cite{rl_policy_gradients}.
\end{itemize}

\subsection{Dynamic Tree Construction via RL Agent}

Our architecture replaces the static, exhaustive tree generation of ProbTree with a dynamic agent-driven process. At each node in the reasoning tree, the system extracts a state representation based on semantic and structural features. This state is passed to an RL agent, which predicts a probability distribution over available reasoning actions: Closed-Book (CB), Open-Book (OB), or Child decomposition. In certain setups, additional actions like Reformulation or Resampling are also included to further enhance the adaptability and expressiveness of the reasoning process. Rather than evaluating all possible actions, the agent samples and executes only the most promising one.

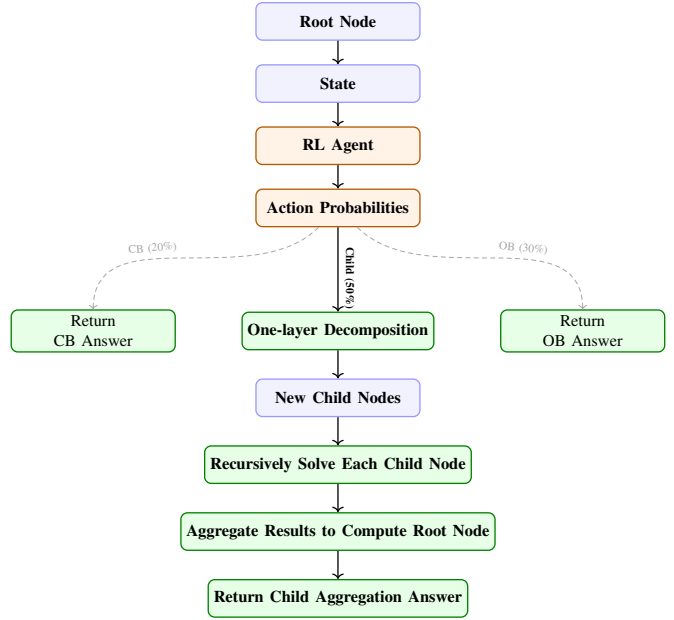
\begin{figure}[t]
\centering
\begin{adjustbox}{max width=0.48\textwidth}
\begin{tikzpicture}[
    datanode/.style={rectangle, draw=blue!40, fill=blue!5, thick, rounded corners, minimum width=3.5cm, minimum height=0.8cm, align=center},
    modelnode/.style={rectangle, draw=orange!70!black, fill=orange!10, thick, rounded corners, minimum width=3.5cm, minimum height=0.8cm, align=center},
    processnode/.style={rectangle, draw=green!50!black, fill=green!10, thick, rounded corners, minimum width=3.5cm, minimum height=0.8cm, align=center},
    dashedarrow/.style={gray!70, densely dashed, ->},
    thickarrow/.style={thick, ->}
]

% Node Definitions
\node[datanode]        (root)                         {\textbf{Root Node}};
\node[datanode, below=0.5cm of root]     (state)           {\textbf{State}};
\node[modelnode, below=0.5cm of state]   (agent)           {\textbf{RL Agent}};
\node[modelnode, below=0.5cm of agent]   (probs)           {\textbf{Action Probabilities}};

\node[processnode, below=1.8cm of probs] (decomp)     {\textbf{One-layer Decomposition}};
\node[processnode, left=1.4cm of decomp] (cb_ans)     {Return\\CB Answer};
\node[processnode, right=1.4cm of decomp] (ob_ans)    {Return\\OB Answer};

\node[datanode, below=0.6cm of decomp]   (children)         {\textbf{New Child Nodes}};
\node[processnode, below=0.6cm of children] (recurse)       {\textbf{Recursively Solve Each Child Node}};
\node[processnode, below=0.6cm of recurse] (agg)            {\textbf{Aggregate Results to Compute Root Node}};
\node[processnode, below=0.6cm of agg]   (return_child)     {\textbf{Return Child Aggregation Answer}};

% Arrows
\draw[thickarrow] (root) -- (state);
\draw[thickarrow] (state) -- (agent);
\draw[thickarrow] (agent) -- (probs);

\draw[thickarrow] (probs) -- (decomp) 
    node[pos=0.6, sloped, above] {\scriptsize \textbf{Child (50\%)}};

\draw[dashedarrow] (probs) to[out=225, in=90] 
    node[pos=0.6, sloped, above, gray!70]{\scriptsize CB (20\%)} (cb_ans.north);

\draw[dashedarrow] (probs) to[out=315, in=90] 
    node[pos=0.6, sloped, above, gray!70]{\scriptsize OB (30\%)} (ob_ans.north);

\draw[thickarrow] (decomp) -- (children);
\draw[thickarrow] (children) -- (recurse);
\draw[thickarrow] (recurse) -- (agg);
\draw[thickarrow] (agg) -- (return_child);

\end{tikzpicture}
\end{adjustbox}
\caption{
\textbf{RL agent-guided reasoning.} The figure illustrates how the agent selects among CB, OB, or Child actions. 
Here, Child (50\%) is chosen, expanding the reasoning tree, while CB (20\%) and OB (30\%) would instead return answers directly. \\
\textbf{Legend:} \textcolor{blue!70!black}{\textbf{Blue}} = Data/Input, 
\textcolor{orange!70!black}{\textbf{Orange}} = Model/Decision, 
\textcolor{green!50!black}{\textbf{Green}} = Reasoning Process.
}
\label{fig:rl_reasoning_flow}
\end{figure}

Figure~\ref{fig:rl_reasoning_flow} illustrates this pipeline. The process begins with a root question node, where the RL agent selects the Child action due to its highest predicted probability. This action triggers a decomposition step that generates new child nodes, which are added to the reasoning tree. Each of these child nodes is then treated as a new decision point, where the agent re-applies the same reasoning strategy recursively.

If instead CB or OB had been selected, the system would have returned the respective answer directly, terminating that branch. This mechanism not only reduces computation by avoiding unnecessary LLM calls but also leads to more targeted, efficient exploration of reasoning paths.

\subsection{Deep Q-Learning Architecture}
We employ Deep Q-Networks (DQN)~\cite{deepqnet} to approximate the optimal policy $\pi^*$ that selects actions maximizing the expected reward. The Q-value function $Q(s,a)$ is learned using experience replay and a target network for stable updates.

\textbf{Reward Function:} Given a chosen answer $a_{\text{chosen}}$, gold answer $a_{\text{gold}}$, and LLM call cost $c$, the reward is computed as:
\begin{equation}
    R = \alpha \cdot \text{sim}(a_{\text{chosen}}, a_{\text{gold}}) - \beta \cdot c
\end{equation}
where $\text{sim}(\cdot)$ is the cosine similarity between sentence embeddings and $\alpha, \beta$ are tunable hyperparameters.

\subsection{Model Variants}
We implemented and evaluated several agent architectures:

\subsubsection{Greedy Hand-Crafted Baseline}

This baseline mimics the modular decision-making structure of ProbTree, but replaces its exhaustive evaluation with a greedy, classifier-driven selection mechanism designed to reduce computation. Rather than executing all reasoning strategies (CB, OB, Child) and selecting the best post-hoc as in ProbTree, we train separate binary classifiers—one for each method—that estimate whether the answer produced by that method is likely to be correct.

\begin{figure}[t]
    \hspace{-0.27cm} % adjust value until it looks good
    \includegraphics[width=0.5\textwidth]{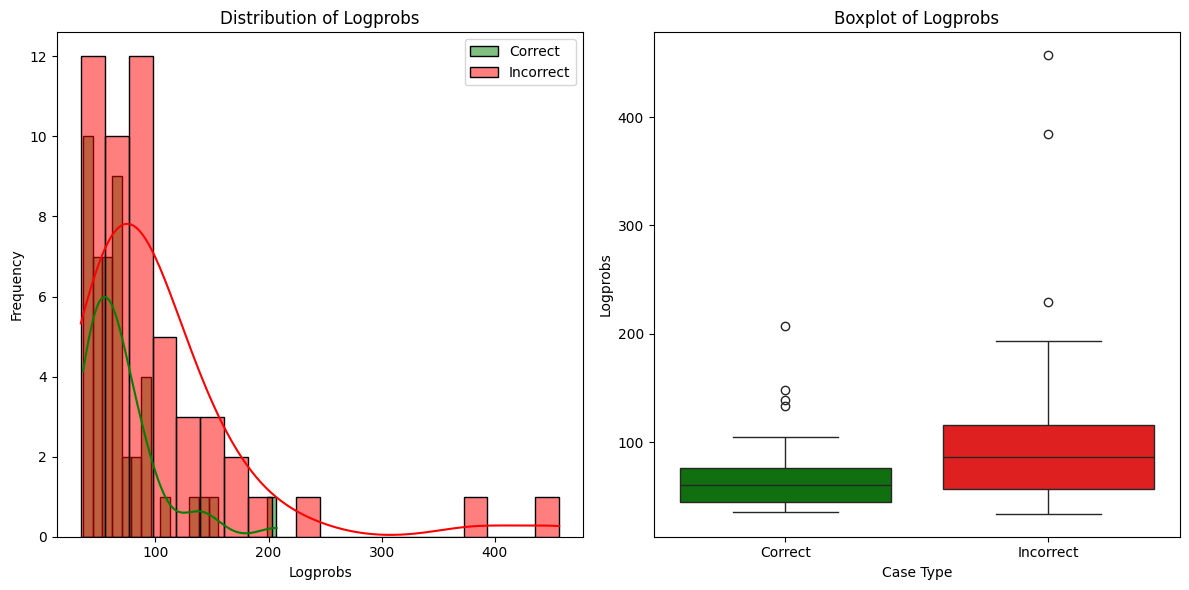}
    \caption{Distribution of log-probabilities for correct vs. incorrect answers using the CB strategy on the dev set.}
    \label{fig:cb_logprobs_corr_incorr}
\end{figure}

These classifiers are Random Forest models trained on log-probability-based features extracted from the development set. For each strategy (CB, OB, Child), we observed a clear correlation between the likelihood of correctness and the average log-probabilities of the generated answer tokens. This pattern is illustrated for CB in Figure~\ref{fig:cb_logprobs_corr_incorr}, and similar trends hold across the other methods.

To exploit this insight efficiently at inference time, we define a fixed evaluation order based on the classifiers' performance on the devsets. For example, in one dataset, the system may evaluate actions in the order CB → OB → Child. The system begins by generating only the CB answer and extracting its log-probabilities. These are passed to the CB classifier. If the classifier predicts that the CB answer is reliable, the system immediately returns it—skipping OB and Child entirely. If the CB prediction is deemed unreliable, the system evaluates OB next, and only evaluates Child if OB is also deemed unreliable.

This sequential, early-stopping design significantly reduces the number of model executions and LLM calls while preserving high decision quality. By combining statistical reliability predictions with a greedy control flow, this approach offers a computationally efficient alternative to both ProbTree and learning-based agents.

\subsubsection{DQN with Question-Only State}
\label{ss: dqn-qonly}
The state includes:
\begin{itemize}
    \item Basic features (has children, question length, number of children, success rates for each action)
    \item Sentence embedding of the question (MiniLM)
    \item Structural depth and position in the tree
\end{itemize}
This compact state is efficient and simulates low-resource reasoning.

\subsubsection{Transformer-based DQN}
Instead of an MLP \cite{mlp}, this variant uses a Transformer encoder to process the same state vector used in \ref{ss: dqn-qonly}, allowing for potential future extensions to sequence-based state tracking.

\subsubsection{DQN with Question + CB + OB}
Extends the previous variant's state by including:
\begin{itemize}
    \item Log-probabilities of CB and OB answers
    \item Sentence embeddings of CB and OB answers
    \item Answer confidence scores and answer lengths
\end{itemize}
This variant provides richer context for decision-making by incorporating features from Closed-Book (CB) and Open-Book (OB) reasoning strategies.

\subsubsection{DQN with Reformulation Actions}
The action space is expanded to:
\texttt{CB, OB, Child, CB\_REFORMULATE, OB\_REFORMULATE, Child\_REFORMULATE.}

The agent can reformulate ambiguous questions before answering, improving robustness and flexibility.

\subsubsection{DQN with Resampling Action}
To address cases where the initial tree decomposition is suboptimal, we introduce a specialized \texttt{RESAMPLE\_CHILDREN} action. This action enables the agent to discard and regenerate the subtree for a given node, effectively giving it a second chance to revise earlier, suboptimal decompositions.

The state is enriched with indicators of subtree uncertainty and confidence variance, allowing the agent to identify when resampling is necessary. The agent is rewarded for correcting earlier failures through effective resampling and penalized for excessive or unnecessary use of this action, ensuring efficient resource allocation.

% For a more comprehensive comparison of state features across these model variants, please see Appendix~\ref{sec:state-representations}, Table~\ref{tab:state_features_usage}.

\subsection{Training Regimes}
Each variant was trained under different reward configurations to capture the trade-off between accuracy and efficiency. We experimented with multiple values of $\alpha$ and $\beta$ on the development sets and found the following settings to be most effective for their intended objectives:

\begin{itemize}
    \item \textbf{High Accuracy}: $\alpha=2.0, \beta=0.05$
    \item \textbf{Balanced}: $\alpha=1.0, \beta=0.1$
    \item \textbf{Efficiency Focused}: $\alpha=0.5, \beta=0.2$
\end{itemize}

\subsection{Summary of Architectures}
Table \ref{table:architectures} summarizes the different reinforcement learning variants discussed in this paper. The table highlights the models, their state complexity, action space, and any additional relevant notes.

\begin{table*}[!htbp] % Use table* for two-column table
\caption{Summary of reinforcement learning variants.}
\centering
\begin{tabular}{|l|c|c|l|}
\hline
\textbf{Model} & \textbf{State Complexity} & \textbf{Action Space} & \textbf{Key Features} \\
\hline
DQN (Q-only)           & Low & CB, OB, Child    & Constrained features \\
Transformer DQN        & Low   & CB, OB, Child    & Encoder-based   \\
DQN (Q+CB+OB)          & High   & CB, OB, Child    & Richer features \\
DQN + Reformulation    & High   & CB, OB, Child (+ Reformulated Variants)    & More expressive \\
DQN + Resampling       & High   & CB, OB, Child, Resample Children    & Tree revision support \\
\hline
\end{tabular}
\label{table:architectures} % Label for referencing
\end{table*}

\section{Experimental Setup}
In this section, we describe our experimental setup, including the retrieval mechanism, evaluation metrics, and training/testing configurations. Our goal is to analyze model performance in terms of accuracy, efficiency, and generalization across multiple multi-hop QA benchmarks.

\subsection{Datasets}
We use the following datasets for training and evaluation:
\begin{itemize}
    \item \textbf{HotpotQA}~\cite{hotpotqa} -- Multi-hop question answering with sentence-level supporting facts.
    \item \textbf{Musique}~\cite{musique} -- Complex multi-hop QA dataset with multiple reasoning steps.
    \item \textbf{2WikiMultiHopQA (2Wiki)}~\cite{2wiki} -- Diverse multi-hop QA over Wikipedia.
\end{itemize}

\subsection{Retriever Details}
For all experiments, we adopt the retrieval component from the original ProbTree implementation~\cite{probtree}. Following IRCoT~\cite{ircot}, we use BM25~\cite{bm25} implemented via Elasticsearch as our retriever. For each node \(q_i \in T\), we retrieve the top-K paragraphs \(R(q_i)\), with \(K \in \{3, 5, 7\}\) chosen based on development set performance. No additional reranking strategies are applied.

\subsection{Evaluation Metrics}
We assess performance using both semantic correctness and efficiency measures:
\begin{itemize}
    \item \textbf{Accuracy}: Percentage of predictions judged as correct by the LLM-as-Judge~\cite{llmasajudge}, where the judge model provides binary (0/1) assessments, capturing diverse valid outputs beyond traditional metrics (e.g., EM, F1).
    \item \textbf{LLM Call Count}: Total number of LLM invocations during inference.
\end{itemize}

\subsection{Training and Testing Configurations}
We evaluate each model under the following configurations:
\begin{enumerate}
    \item \textbf{In-domain Evaluation}: Train and test on the same dataset (e.g., HotpotQA $\rightarrow$ HotpotQA).
    \item \textbf{Cross-domain Generalization}: Train on one dataset and test on the others (e.g., HotpotQA $\rightarrow$ Musique, 2Wiki).
    \item \textbf{Multi-domain Generalization}: Train on all datasets and test across all splits.
\end{enumerate}

\section{Evaluation and Results}

\subsection{Baselines and Comparison Scope}

We primarily compare our proposed models against the original ProbTree framework, which was already evaluated against a range of adaptive reasoning baselines, including IRCoT~\cite{ircot} and ReAct~\cite{react}, among others, using the same datasets, retrievers, and evaluation settings~\cite{probtree}. Therefore, demonstrating improvements over ProbTree on our datasets implies improvements over these baselines as well. For clarity and reproducibility, we refer readers to the original ProbTree paper for detailed comparisons with these adaptive methods.

\subsection{LLM Setup}
Our experiments were conducted using the LLM \texttt{meta-llama/Llama-3.3\allowbreak-70B\allowbreak-Instruct\allowbreak-Turbo\allowbreak-Free}. For interaction with the LLM, we utilized the `Together.ai` API with the following configuration: 
\texttt{max\_tokens=None}, \texttt{stop=None}, and \texttt{use\_cache=True}. Unless otherwise specified, we used a temperature of 0 for all queries in the normal setup, as this setting ensures deterministic responses by minimizing randomness.

For the \textbf{resampling} and \textbf{reformulation} variants, we adjusted the settings to specifically handle the need for diverse outputs. These evaluations were conducted with a higher temperature of 0.7 and with caching disabled to encourage the generation of alternative formulations or varied responses during reasoning. Such adjustments ensured meaningful experimentation for these strategies, where generating diverse outputs is critical for performance.

\noindent For reproducibility, we adopt the same prompt templates as ProbTree~\cite{probtree}. The exact prompts and additional qualitative examples are available in our project repository.

\subsection{In-domain Results}
\subsubsection{In-domain Accuracy Comparison Across Datasets}

We consolidate the in-domain evaluation results across the three datasets—HotpotQA, 2Wiki, and Musique—by including each model's performance under all three reward configurations. Table~\ref{tab:all-datasets-accuracy} presents a comprehensive view of how different models adapt across datasets and reward priorities.

\begin{table*}[h!]
\centering
\compacttables  
\caption{In-domain accuracy (\%) across reward configurations and datasets. 
"Orig." denotes models in their original setup (e.g., ProbTree, Greedy, Random) without reward configurations. Other columns report RL-based models under High, Balanced, and Efficient rewards.}
\setlength{\tabcolsep}{3pt} % Adjust column spacing
\renewcommand{\arraystretch}{1.2} % Adjust row spacing
\begin{tabular}{|l|cccc|cccc|cccc|}
\hline
\multirow{2}{*}{\textbf{Model}} & \multicolumn{4}{c|}{\textbf{HotpotQA}} & \multicolumn{4}{c|}{\textbf{2Wiki}} & \multicolumn{4}{c|}{\textbf{Musique}} \\ 
                                & Orig.       & High     & Bal.      & Eff.      & Orig.       & High      & Bal.      & Eff.      & Orig.      & High      & Bal.      & Eff.      \\ \hline
Original ProbTree               & \textbf{71.40} & -      & -      & -       & 74.60       & -      & -      & -       & \textbf{44.80}       & -      & -      & -       \\ 
Greedy Solver                   & 70.40       & -      & -      & -       & \textbf{77.00} & -      & -      & -       & 44.20 & -      & -      & -       \\ 
Random Agent                    & 62.40       & -      & -      & -       & 59.80       & -      & -      & -       & 31.40       & -      & -      & -       \\ \hline
DQN (Q-only)                    & -         & 58.60    & 58.80    & 60.40     & -         & 57.00    & 55.20    & 53.40     & -       & 26.60    & 26.00    & 26.60     \\ 
Transformer DQN                 & -         & 69.20 & \textbf{62.20} & \textbf{61.00} & -    & 48.80    & 57.00    & 48.80     & -       & 24.80    & 24.20    & 26.80     \\ 
DQN (Q+CB+OB)                   & -         & 61.20    & 58.20    & 60.60     & -         & 54.20    & 54.40    & 53.00     & -       & 26.20    & 27.40    & 26.20     \\ 
DQN + Reform.                   & -         & 56.20    & 58.80    & 60.20     & -         & 56.40    & 53.60    & \textbf{56.20} & -    & 26.40    & 25.00    & 27.00     \\ 
DQN + Resampling                & -         & \textbf{69.60}    & 52.20    & 59.40     & -         & \textbf{62.80}    & \textbf{75.60} & 52.60 & -       & \textbf{28.20}    & \textbf{45.00} & \textbf{28.60} \\ 
\hline
\end{tabular}
\label{tab:all-datasets-accuracy}
\end{table*}

The results in Table~\ref{tab:all-datasets-accuracy} provide a unified comparison of in-domain performance across models and datasets. While static baselines like ProbTree and the Greedy solver remain strong on HotpotQA and 2Wiki, they underperform significantly on Musique, suggesting that dataset structure plays a critical role in reasoning difficulty.

Among the dynamic RL agents, the DQN with resampling demonstrates notable adaptability across all datasets. It achieves the best result on HotpotQA (69.60\%), 2Wiki (75.60\%), and also obtains the top score on Musique under the Balanced configuration (45.00\%)—surpassing both static and learned alternatives. This highlights the value of allowing dynamic decomposition revisions when initial plans are suboptimal.

The Transformer-based agent performs best on HotpotQA but struggles to generalize well to the Musique dataset, where even the Random policy outperforms many of the RL variants. This indicates potential issues with overfitting or limited expressivity in structurally sparse contexts.

Overall, results across configurations reinforce the importance of dataset characteristics, model design, and reward shaping in dynamic tree-of-thought reasoning tasks. Future analyses will examine efficiency and method usage patterns in more detail.

\subsubsection{Accuracy-Cost Tradeoff Analysis}

\begin{figure*}[t]
    \centering
    \includegraphics[width=\textwidth,keepaspectratio]{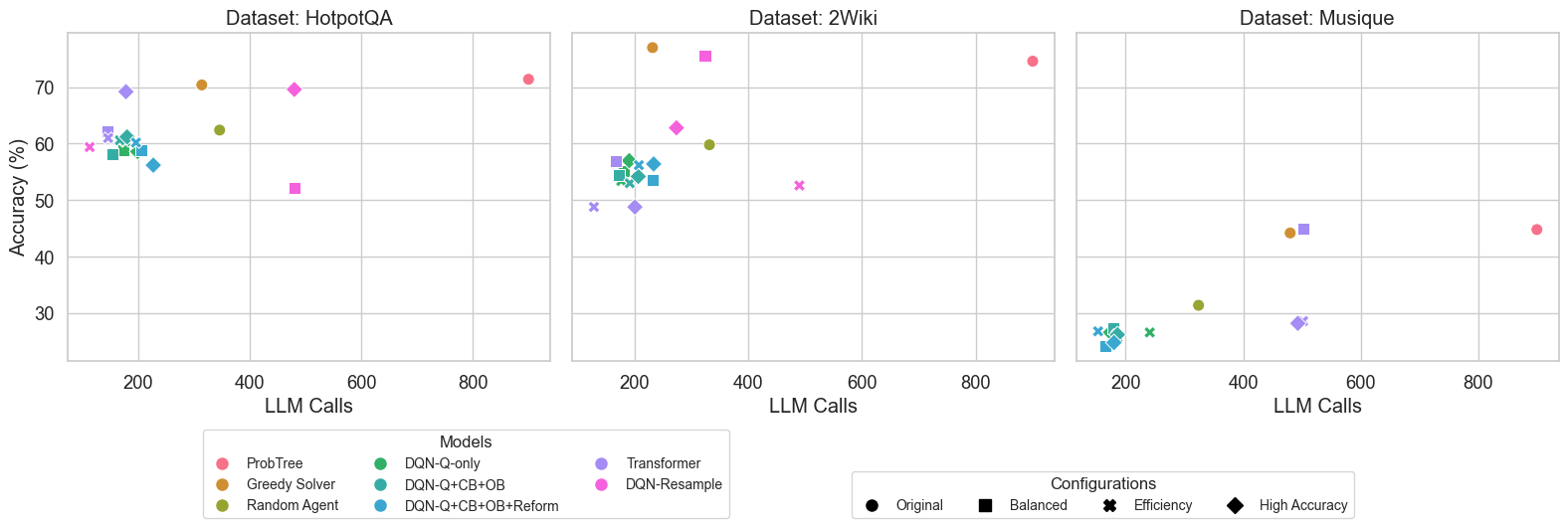}
    \caption{Accuracy vs. Number of LLM calls tradeoff across datasets, models, and configurations. This evaluation was conducted on a subset of 100 test examples per dataset, specifically for analyzing efficiency.}
    \label{fig:acc_vs_cost_scatter}
\end{figure*}

Balancing accuracy and computational cost is crucial for real-world deployment, where LLM calls dominate the cost. Table~\ref{tab:all-datasets-accuracy} shows individual accuracy metrics, but Figure~\ref{fig:acc_vs_cost_scatter} visualizes the tradeoff across datasets. Each point represents a (model, configuration, dataset) tuple, with LLM calls on the x-axis and accuracy on the y-axis. Models are differentiated by color, configurations by marker shapes, and datasets by subplot groups.

Key insights from the analysis are:
\begin{itemize}
    \item \textbf{Resampling (2Wiki and Musique):}  
    DQN-Resample achieves the best tradeoff in the Balanced setting, delivering \textbf{75.6\% accuracy with 324 calls on 2Wiki} and \textbf{45.0\% accuracy with 503 calls on Musique}, outperforming static baselines like ProbTree.

    \item \textbf{Transformer Efficiency (HotpotQA):}  
    Transformer agents excel at balancing accuracy and cost on HotpotQA, achieving \textbf{62.2\% accuracy with 144 calls (Balanced)} and \textbf{61.0\% accuracy with 146 calls (Efficiency)} while significantly reducing cost compared to ProbTree or Greedy Solver.

    \item \textbf{Reformulation Costs:}  
    Reform-based models incur higher costs (e.g., \textbf{196–227 calls on HotpotQA}) but provide only marginal accuracy improvements, making simpler configurations like Balanced more cost-effective.

    \item \textbf{Static Baselines:}  
    ProbTree consistently incurs the highest cost (\textbf{900 calls}) while achieving competitive accuracy on HotpotQA (\textbf{71.4\%}) and 2Wiki (74.6\%) but underperforming on Musique (44.8\%).  
    In contrast, Greedy Solver is substantially more cost-efficient: it outperforms all methods on 2Wiki (\textbf{77.0\%}), remains competitive on HotpotQA (70.4\%), but lags behind on Musique (44.2\%).  
    Random Agent performs the worst across all datasets.

\end{itemize}

This analysis highlights the strength of learned agents (e.g., Transformer and DQN-Resample) in optimizing the accuracy-cost tradeoff, while static baselines and reform-based approaches are less efficient under cost constraints.

\subsection{Results: Cross-Domain Generalization}

To evaluate the generalizability of our RL-based approaches, we conducted experiments where models trained on one dataset were tested on other unseen datasets, analyzing their performance and adaptability. Table~\ref{tab:all-datasets-accuracy} summarizes the in-domain accuracy, serving as a benchmark, while Table~\ref{tab:cross-domain-summary} highlights the best accuracy observed for each cross-domain train-test pair.

From these results, several key insights can be drawn:

\begin{itemize}
    \item \textbf{Greedy Provides the Strongest Baseline:}  
    The Greedy configuration consistently achieves the highest or near-highest cross-domain accuracy across train–test scenarios, establishing itself as the most reliable static baseline. 
    For example, it reaches 69.20\% for Musique→HotpotQA and 73.00\% for Musique→2Wiki. 
    
    \item \textbf{Drop in Cross-Domain Accuracy:}  
    Cross-domain accuracy is generally lower than in-domain accuracy. For instance, training on Musique and testing on HotpotQA yields 63.00\% accuracy (Transformer, Efficiency), which is approximately 7.4\% lower compared to HotpotQA's in-domain benchmark of 70.40\%. This highlights the challenges of generalization, as models often fall short of replicating in-domain performance when applied to unseen datasets.

    \item \textbf{Impact of Training Dataset:}  
    The choice of training dataset significantly affects generalization. Models trained on 2Wiki perform better on HotpotQA (63.40\%) compared to Musique (36.60\%). Similarly, training on Musique leads to strong transfer to 2Wiki (65.20\%), but less so to HotpotQA (63.00\%).

    \item \textbf{Resampling and Transformer Strengths:}  
    Resampling-based models consistently achieve competitive cross-domain results, such as 65.20\% for Musique → 2Wiki (Efficiency) and 63.40\% for 2Wiki → HotpotQA (High Accuracy). Transformers also generalize well, particularly for Musique → HotpotQA (63.00\%, Efficiency).

    \item \textbf{Challenges in Transfer to Musique:}  
    Musique consistently poses a challenge for cross-domain generalization, with accuracy struggling to exceed 36.60\%. This trend is noticeable even in in-domain results, where Musique achieves the lowest benchmark across all configurations.

    \item \textbf{Efficiency Configurations Perform Well:}  
    Efficiency-focused configurations deliver strong generalization, striking a balance between accuracy and resource usage across most train-test setups.
\end{itemize}

In summary, while cross-domain accuracy is generally lower than in-domain performance, certain approaches—such as Resampling and Transformers—exhibit strong adaptability. However, transferring to Musique remains challenging, likely due to its domain-specific complexities.

\begin{table*}[h!]
\centering
\caption{Best generalization results across train–test pairs. 
For each test dataset, we report: (1) its best in-domain accuracy (upper bound), and for each train–test pair we provide two cross-domain results: (a) the Greedy Solver in its original configuration, serving as the strongest static baseline, and (b) the best-performing RL-based model under any reward configuration.}
\label{tab:cross-domain-summary}
\setlength{\tabcolsep}{6.5pt} % Adjust column spacing
\renewcommand{\arraystretch}{1.3} % Adjust row spacing
\compacttables %TODO:: compact
\begin{tabular}{|>{\centering\arraybackslash}m{2.2cm}|
                >{\centering\arraybackslash}m{2.2cm}|
                >{\centering\arraybackslash}m{2cm}|
                >{\centering\arraybackslash}m{2cm}|
                >{\centering\arraybackslash}m{2cm}|
                >{\centering\arraybackslash}m{3cm}|}
\hline
\textbf{Train Dataset} & \textbf{Test Dataset} & \textbf{Best In-Domain Accuracy (\%)} & \textbf{Best Cross-domain Accuracy (\%)} & \textbf{Model} & \textbf{Configuration} \\
\hline
\multirow{2}{*}{Musique} & \multirow{2}{*}{HotpotQA} & \multirow{2}{*}{70.40} 
& \textbf{69.20} & Greedy & Original \\ 
\cline{4-6}
 &  &  & 63.00 & Transformer & Efficiency \\
\hline
\multirow{2}{*}{Musique} & \multirow{2}{*}{2Wiki} & \multirow{2}{*}{77.00} 
& \textbf{73.00} & Greedy & Original \\ 
\cline{4-6}
 &  &  & 65.20 & Resampling & Efficiency \\
\hline
\multirow{2}{*}{2Wiki} & \multirow{2}{*}{Musique} & \multirow{2}{*}{45.00} 
& \textbf{46.00} & Greedy & Original \\ 
\cline{4-6}
 &  &  & 36.60 & Resampling & Balanced \\
\hline
\multirow{2}{*}{2Wiki} & \multirow{2}{*}{HotpotQA} & \multirow{2}{*}{70.40} 
& \textbf{70.60} & Greedy & Original \\ 
\cline{4-6}
 &  &  & 63.40 & Resampling & High Accuracy \\
\hline
\multirow{2}{*}{HotpotQA} & \multirow{2}{*}{2Wiki} & \multirow{2}{*}{77.00} 
& \textbf{74.40} & Greedy & Original \\ 
\cline{4-6}
 &  &  & 71.60 & Transformer & High Accuracy \\
\hline
\multirow{2}{*}{HotpotQA} & \multirow{2}{*}{Musique} & \multirow{2}{*}{45.00} 
& \textbf{44.20} & Greedy & Original \\ 
\cline{4-6}
 &  &  & 36.60 & Resampling & Balanced \\
\hline
\end{tabular}
\end{table*}

\subsection{Results: Multi-Domain Generalization}

Table~\ref{tab:multi_domain_results} summarizes the results for models trained on all three datasets and tested individually on HotpotQA, 2Wiki, and Musique. The performance of the models is evaluated across various RL-based approaches and configurations, as well as baselines such as the Original ProbTree and Greedy Solver.

\begin{table*}[h!]
\centering
\caption{Multi-Domain Generalization Results: Accuracy (\%) across Datasets and Approaches.}
\label{tab:multi_domain_results}
\compacttables
\begin{tabular}{|>{\centering\arraybackslash}m{5.1cm}|
                >{\centering\arraybackslash}m{3cm}|
                >{\centering\arraybackslash}m{2cm}|
                >{\centering\arraybackslash}m{2cm}|
                >{\centering\arraybackslash}m{2cm}|}
\hline
\textbf{Approach} & \textbf{Configuration} & \textbf{HotpotQA} & \textbf{2Wiki} & \textbf{Musique} \\
\hline
Original ProbTree & - & \textbf{71.40} & \textbf{74.60} & \textbf{44.80} \\ \hline
Greedy Solver     & - & \textbf{70.80} & \textbf{77.20} & \textbf{45.20} \\ \hline

\multirow{3}{*}{RL (Question Only)} 
& High Accuracy       & 58.60 & 55.20 & 25.20 \\
& Balanced            & 61.00 & 57.80 & 26.60 \\
& Efficiency Focused  & 58.00 & 55.00 & 26.40 \\ \hline

\multirow{3}{*}{RL (Question + CB + OB)} 
& High Accuracy       & 60.00 & 55.20 & 24.80 \\
& Balanced            & 57.40 & 53.20 & 25.00 \\
& Efficiency Focused  & 61.80 & 55.40 & 26.60 \\ \hline

\multirow{3}{*}{RL (Question + CB + OB + Reformulation)} 
& High Accuracy       & 61.80 & 58.80 & 25.80 \\
& Balanced            & 58.40 & 55.40 & 25.80 \\
& Efficiency Focused  & 59.60 & 52.40 & 26.80 \\ \hline

\multirow{3}{*}{RL (Transformer, Question Only)} 
& High Accuracy       & \textbf{69.20} & 71.60 & 21.60 \\
& Balanced            & 61.00 & 53.00 & 27.80 \\
& Efficiency Focused  & 62.00 & 56.80 & 26.60 \\ \hline

\multirow{3}{*}{RL (Resampling)} 
& High Accuracy       & 67.80 & 74.00 & \textbf{41.60} \\
& Balanced            & 67.80 & \textbf{77.60} & 32.40 \\
& Efficiency Focused  & 66.00 & 69.00 & 38.20 \\ \hline
\end{tabular}
\end{table*}

On 2Wiki, resampling achieves the highest accuracy at 77.60\% (Balanced configuration), marginally surpassing the in-domain best result of 75.60\%, as shown in Table~\ref{tab:all-datasets-accuracy}. This highlights the ability of resampling to generalize effectively under the multi-domain setup.

For HotpotQA, the best performance in the multi-domain setup, 69.20\% (High Accuracy configuration using Transformers), matches the top accuracy observed in in-domain training. This result demonstrates robust generalization across multiple approaches for this dataset.

On Musique, results are slightly lower compared to in-domain performance. The top-performing model in the multi-domain setup, resampling with 41.60\% accuracy, falls slightly short of the in-domain best result, which stands at 45.00\% (Resampling). This reflects the inherent challenges of achieving strong generalization on this dataset.

Overall, resampling consistently produces competitive results across the datasets and achieves significant improvements on 2Wiki. Transformer-based RL models also show strong generalization capabilities, particularly for textual datasets like HotpotQA and 2Wiki. Despite minor performance drops for Musique, the multi-domain setup demonstrates its ability to retain high performance across challenging datasets.

\subsection{Method Usage Analysis}

To better understand how different configurations influence the reasoning strategies employed by our models, we analyze method usage statistics across all datasets. Figure~\ref{fig:method_counts_config} presents the frequency of each method selected under the three reward configurations: High Accuracy, Balanced, and Efficiency Focused.
Several clear trends emerge:
\begin{figure*}[t]
    \centering
    \includegraphics[width=\textwidth]{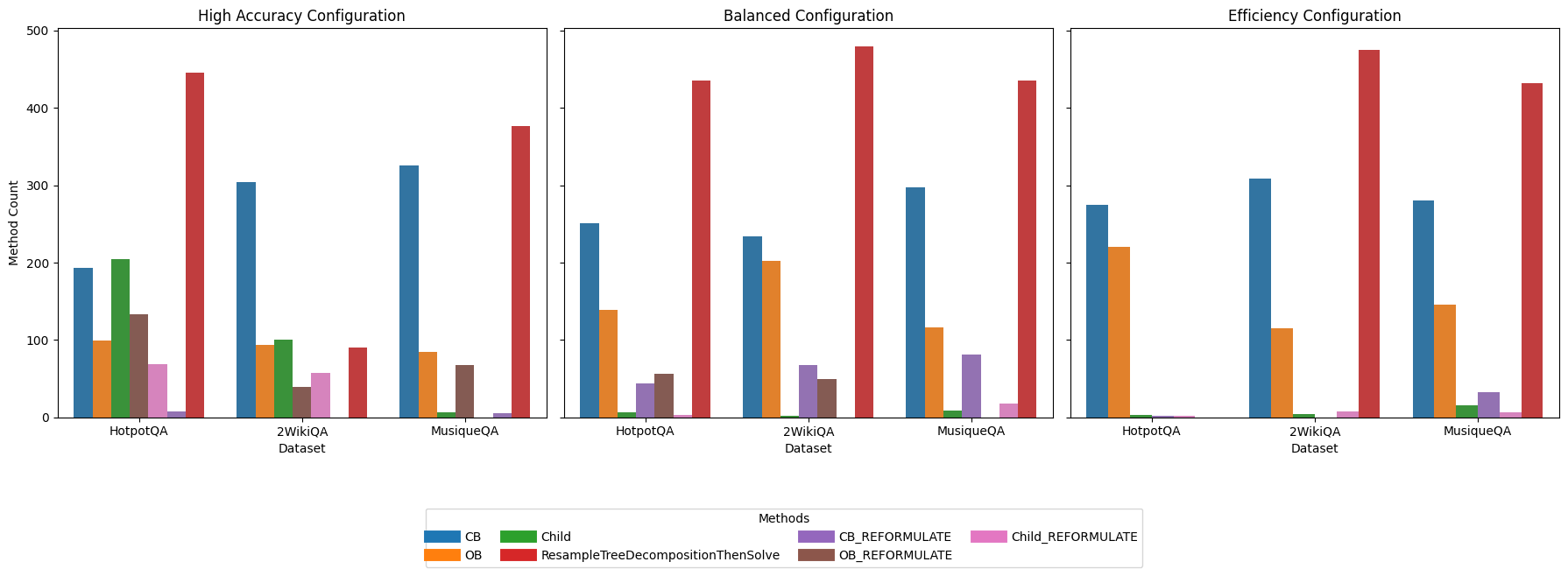}
    \caption{Distribution of selected reasoning strategies across reward configurations and datasets. CB = Closed-Book, OB = Open-Book, Child = Decomposition, and Reformulation/Resampling variants.}
    \label{fig:method_counts_config}
\end{figure*}

\begin{itemize}
    \item \textbf{High Accuracy configuration} tends to favor deeper decompositions and reformulations, particularly in datasets like HotpotQA and Musique, where the Child and Resample actions are used frequently. This confirms the agent’s inclination toward more exploratory reasoning paths when accuracy is prioritized.
    
    \item \textbf{Balanced configuration} demonstrates a hybrid behavior. The ResampleTreeDecompositionThenSolve method becomes dominant in most datasets, indicating the agent often revisits and improves initial decompositions while still maintaining cost-awareness. CB and OB are still used selectively, showing that the model is learning to allocate resources strategically.
    
    \item \textbf{Efficiency Focused configuration} significantly reduces reliance on decomposition. Instead, CB and OB become the dominant strategies, especially in datasets like 2Wiki and HotpotQA. This highlights how cost penalties effectively steer the agent toward quicker, single-shot answering mechanisms.
\end{itemize}

This analysis illustrates that the RL agent is successfully adapting its decision-making policies to align with the given reward priorities. The modular action space combined with learning-based selection enables the agent to strike a nuanced balance between depth and efficiency in multi-hop reasoning.

% Additional breakdowns by model variant are provided in Appendix~\ref{appendix:method-usage-models}.

\section{Related Work}
\subsection{Structured Prompting and Reasoning Topologies}

Recent advancements in structured prompting have introduced various reasoning topologies to enhance the capabilities of LLMs. Besta et al.~\cite{demystifyingchainstreesgraphs} provide a comprehensive taxonomy of these strategies, categorizing them along three dimensions: \textit{Topology Scheme}, \textit{Reasoning Schedule}, and \textit{AI Pipeline} involvement.

\textbf{Topology Scheme}: This dimension classifies the structural format of reasoning, such as chains~\cite{cot}, trees~\cite{tot}, or graphs. Our approach aligns with the \textit{Tree-of-Thoughts}~\cite{tot} paradigm, where reasoning is structured hierarchically. However, unlike static trees, our method employs a dynamic tree construction guided by a RL agent, allowing for adaptive reasoning paths based on the problem context.

\textbf{Reasoning Schedule}: This refers to the sequence and manner in which reasoning steps are executed. Traditional methods often follow a predetermined schedule. In contrast, our RL-based framework dynamically determines the reasoning path, selecting actions like Closed-Book, Open-Book, or decomposition based on learned policies that balance accuracy and computational efficiency.

\textbf{AI Pipeline}: This aspect considers the components of the AI system utilized during reasoning. While many existing methods rely solely on prompting, our approach integrates additional components such as retrieval mechanisms and tool usage, enhancing the model's ability to handle complex, knowledge-intensive tasks.

\subsection{Adaptive Reasoning Frameworks}

Building upon structured prompting, adaptive reasoning frameworks have been proposed to allow LLMs to adjust their reasoning strategies dynamically. Pandey et al.~\cite{pandey2025adaptive} introduce the Adaptive Graph of Thoughts (AGoT) framework, which constructs a dynamic directed acyclic graph (DAG) of interdependent reasoning steps at test time. AGoT recursively decomposes complex queries into structured subproblems, selectively expanding only those that require further analysis, thereby unifying the strengths of chain, tree, and graph paradigms into a cohesive framework that allocates computation where it is most needed.

Our RL-based approach shares the goal of adaptive reasoning but differs in its methodology. While AGoT operates through recursive decomposition at test time, our framework employs reinforcement learning to learn optimal policies for action selection during training, enabling the model to make informed decisions about when to decompose, retrieve information, or answer directly, thus optimizing both performance and computational efficiency.

% TODO:: remove it ? 
% \subsection{Self-Guided Multimodal Reasoning}

% In the realm of multimodal reasoning, self-guided strategies have been explored to enhance LLM performance. Hu et al.~\cite{socratic} propose the Socratic Questioning framework, where the model engages in self-questioning to guide its reasoning process, particularly in multimodal contexts. This approach heuristically guides multimodal LLMs to focus on relevant visual clues, reducing hallucinations and enhancing the model's ability to describe fine-grained image details.

% While Socratic Questioning emphasizes recursive self-inquiry to improve reasoning depth, our method focuses on strategic action selection through reinforcement learning to optimize both reasoning quality and resource utilization. Both approaches aim to enhance the reasoning capabilities of LLMs, but they operate in different modalities and utilize distinct mechanisms for guiding the reasoning process.

\subsection{Positioning Our Approach}

Integrating insights from structured prompting, adaptive reasoning, and self-guided strategies, our RL-based dynamic Tree-of-Thought framework represents a novel approach to structured reasoning in LLMs. By learning to select among various reasoning strategies based on the current state, our model dynamically constructs reasoning trees that balance depth and efficiency, adapting to the complexity of the task at hand. This approach not only enhances performance on knowledge-intensive tasks but also optimizes computational resources, paving the way for more scalable and effective LLM applications.

\section{Conclusion}

We presented a reinforcement learning-based framework for dynamic Tree-of-Thought (ToT) reasoning that enables LLMs to perform adaptive, multi-step decision-making during inference. By formulating reasoning as a Markov Decision Process, the framework allows the agent to strategically select actions such as decomposition, retrieval, or direct answering, depending on the task's complexity and context. This dynamic construction of reasoning trees balances efficiency and depth, offering significant improvements in both accuracy and computational resource utilization.

Our approach outperforms static prompting strategies by providing a flexible and resource-aware reasoning mechanism aligned with real-world decision-making needs. Additionally, the pruning of redundant computation paths highlights the potential for efficient, knowledge-intensive reasoning.

\textbf{Limitations and Future Work:} Despite the promising results, our framework has certain limitations. While we have introduced new actions, such as reformulation and resampling, that expand the action space beyond traditional reasoning strategies (decomposition, direct answering, retrieval), the framework still primarily operates within a predefined set of reasoning mechanisms. Exploring more semantically complex or task-specific actions represents an important direction for future work, allowing the framework to adapt to a broader range of problem types. Second, generalization across unseen test domains remains an open challenge, particularly for datasets with significant domain or knowledge-specific shifts, such as Musique. Lastly, the interpretability of learned policies could benefit from further refinement to promote greater safety and transparency in reasoning processes.

Future work will explore integrating this framework with fine-tuned multi-agent systems and expanding its applicability to multi-modal tasks requiring both language and visual reasoning. We also envision advancing its use for safe and interpretable AI by further enhancing the explicitness of reasoning trees and making optimization more explainable.

\bibliographystyle{IEEEtran}
\bibliography{custom}

\appendix
\subsection{Complete Cross-Domain Generalization Results}
\label{appendix-full-cross-results}

Table~\ref{tab:appendix-cross-domain-results} presents the full results across all configurations and feature sets for cross-domain generalization experiments. Since the Greedy configuration achieves the best performance across all cases, we highlight it in bold. Additionally, for comparison, we also bold the second-best result among the RL-based configurations (excluding Greedy) for each train-test pair to indicate the next best-performing model in terms of accuracy.

\begin{table*}[h!]
\centering
\caption{Detailed cross-domain generalization results: Accuracy (\%) across train-test pairs. Rows include all feature sets and configurations.}
\compacttables
\label{tab:appendix-cross-domain-results}
\setlength{\tabcolsep}{4pt} % Adjust column spacing
\renewcommand{\arraystretch}{1.2} % Adjust row spacing
\begin{tabular}{|l|c|c|c|c|c|c|}
\hline
\multirow{2}{*}{\textbf{Model + Config}} & \multicolumn{6}{c|}{\textbf{Train → Test Datasets}}                                                                                 \\ \cline{2-7} 
                                         & \textbf{Musique}        & \textbf{Musique}        & \textbf{2Wiki}          & \textbf{2Wiki}          & \textbf{HotpotQA}      & \textbf{HotpotQA}      \\  
                                         & \textbf{HotpotQA}       & \textbf{2Wiki}          & \textbf{Musique}        & \textbf{HotpotQA}       & \textbf{2Wiki}         & \textbf{Musique}       \\ \hline
Greedy  & \textbf{69.20}                   & \textbf{73.00}                   & \textbf{46.00}                   & \textbf{70.60}                   & \textbf{74.40}        & \textbf{44.20}                 \\ \Xhline{1.2pt}
Q-Only (High Accuracy)                  & 59.20                   & 59.20                   & 24.80                   & 59.40                   & 49.00        & 23.20        \\ \hline
Q-Only (Balanced)                       & 60.60                   & 60.60                   & 26.00                   & 59.40                   & 49.00                  & 24.20                 \\ \hline
Q-Only (Efficiency)                     & 61.00                   & 61.00                   & 24.80                   & 56.60                   & 46.80                  & 27.00                 \\ \hline
Q+CB+OB (High Accuracy)                 & 60.80                   & 52.80                   & 25.60                   & 59.60                   & 56.20                  & 26.60                 \\ \hline
Q+CB+OB (Balanced)                      & 61.20                   & 55.00                   & 25.40                   & 60.40                   & 57.60                  & 36.00        \\ \hline
Q+CB+OB (Efficiency)                    & 61.80          & 56.20                   & 27.20          & 57.40                   & 57.60                  & 36.60                 \\ \hline
Transformers DQN (High Accuracy)              & 59.20                   & 52.20                   & 23.20                   & 55.80          & \textbf{71.60}                  & 21.60                 \\ \hline
Transformers DQN (Balanced)              & 57.60                   & 52.80                   & 26.80                   & 62.20          & 57.00                  & 27.00                 \\ \hline
Transformers DQN (Efficiency)              & \textbf{63.00}                   & 57.00                   & 23.20                   & 55.80          & 54.00                  & 27.20                 \\ \hline
Reformulation (High Accuracy)              & 61.20                   & 54.40                   & 25.00                   & 58.40          & 57.60                  & 24.40                 \\ \hline
Reformulation (Balanced)              & 58.60                   & 55.20                   & 26.20                   & 59.80          & 53.40                  & 24.60                 \\ \hline
Reformulation (Efficiency)              & 59.80                   & 57.00                   & 27.20                   & 60.40          & 53.80                  & 27.00                 \\ \hline
Resampling (High Accuracy)              & 57.20                   & 62.40                   & 32.20          & \textbf{63.40}          & 63.40        & 34.40                 \\ \hline
Resampling (Balanced)                   & 60.80          & 60.40                   & \textbf{36.60}                   & 59.40                   & 62.00                  & \textbf{36.60}        \\ \hline
Resampling (Efficiency)                 & 58.60                   & \textbf{65.20}          & 34.00                   & 58.80                   & 55.80                  & 36.40                 \\ \hline
\end{tabular}
\end{table*}

\subsection{Method Usage by Model Variant}
\label{appendix:method-usage-models}

\begin{figure*}[h]
    \centering
    \includegraphics[width=\textwidth]{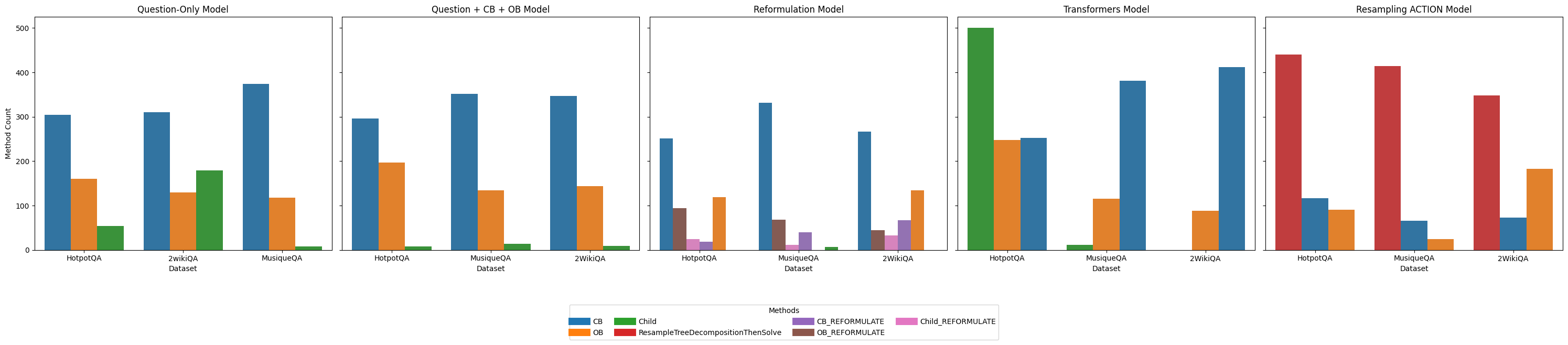}
    \caption{Method usage distribution per model variant across datasets. Each bar reflects the number of times a reasoning method (CB, OB, Child, etc.) was selected.}
    \label{fig:method_counts_model}
\end{figure*}

While the main paper discusses method usage across reward configurations, we complement that view by analyzing method preferences by model architecture. Figure~\ref{fig:method_counts_model} shows the distribution of selected reasoning strategies for each model variant across all datasets.

The breakdown reveals several key observations:

\begin{itemize}
\item \textbf{Question-Only agents} rely heavily on CB and OB, with occasional Child selections, defaulting to simpler strategies.
\item \textbf{Question + CB + OB models} show a similar pattern but with slightly more Child usage from improved state representations derived from answer log-probabilities.
\item \textbf{Reformulation-based agents} display a more balanced distribution across original and reformulated variants, with CB\_REFORMULATE and OB\_REFORMULATE used significantly more, indicating that the additional action space allows for more nuanced strategies.
\item \textbf{Transformer-based models} strongly prefer Child, especially in HotpotQA, aligning with their superior accuracy in that domain.
\item \textbf{Resampling models} are skewed toward the ResampleTreeDecompositionThenSolve action, confirming their tendency to revise poor decompositions to enhance performance.
\end{itemize}

This model-wise analysis complements the configuration-wise view and highlights how architectural choices shape the agent’s use of its action space.

\subsection{State Representations for Models}
\label{sec:state-representations}
Table \ref{tab:state_features_usage} summarizes the features used to construct the state representations for each of our models. These state features include statistical attributes, semantic embeddings, structural information, and confidence scores. Each model includes a subset of the available features based on its specific design and configuration (e.g., Question Only, CB+OB, Resampling, etc.).
\noindent
\textbf{Feature Descriptions.} The state features capture structural, semantic, and statistical aspects of reasoning. Structural features include \textit{Has Children}, \textit{Number of Children}, and \textit{Tree Depth and Position}, reflecting the node’s role and placement in the reasoning tree. Semantic features, such as \textit{Question Embeddings} and \textit{Answer Embeddings}, encode the meaning of inputs and candidate answers for generalization and consistency checks. Statistical and confidence-based features include \textit{Success Rates for Each Method}, \textit{Log Probabilities}, \textit{Answer Lengths}, and \textit{Answer Confidence Scores}, guiding policy decisions and pruning. Finally, \textit{Child-Specific Features} provide per-child versions of key signals (e.g., length, confidence, overlap), enabling fine-grained resampling and prioritization.

\begin{table*}[tb]
\centering
\caption{State Feature Usage Across Models}
\compacttables
\label{tab:state_features_usage}
\begin{tabular}{|l|c|c|c|c|c|}
\hline
\textbf{State Features}          & \textbf{Question Only} & \textbf{Transformers b} & \textbf{CB+OB} & \textbf{Reformulation} & \textbf{Resampling} \\ \hline
Has Children (0/1)                     & \checkmark             & \checkmark            & \checkmark     & \checkmark            & \checkmark          \\ \hline
Question Length and Type          & \checkmark             & \checkmark            & \checkmark     & \checkmark            & \checkmark          \\ \hline
Number of Children               & \checkmark             & \checkmark            & \checkmark     & \checkmark            & \checkmark          \\ \hline
Success Rates for each method                    & \checkmark             & \checkmark            & \checkmark     & \checkmark            & \checkmark          \\ \hline
Question Embedding               & \checkmark             & \checkmark            & \checkmark     & \checkmark            & \checkmark          \\ \hline
Tree Depth and Position          & \checkmark             & \checkmark            & \checkmark     & \checkmark            & \checkmark          \\ \hline
LogProbabilities                 &                        &                        & \checkmark     & \checkmark            & \checkmark          \\ \hline
Answer Embeddings                &                        &                        & \checkmark     & \checkmark            & \checkmark          \\ \hline
Answer Lengths                   &              &             & \checkmark     & \checkmark            & \checkmark          \\ \hline
Answers Confidence Scores                &              &             & \checkmark     & \checkmark            & \checkmark          \\ \hline
Child-Specific Features          &                        &                        &                &                        & \checkmark          \\ \hline
\end{tabular}
\end{table*}

\end{document}